 \documentclass[pmlr,twocolumn,10pt]{jmlr} 

\usepackage{booktabs}
\usepackage{siunitx}

\usepackage[switch]{lineno}

\theorembodyfont{\upshape}
\theoremheaderfont{\scshape}
\theorempostheader{:}
\theoremsep{\newline}

\usepackage[skip=2pt]{caption}   
\usepackage{booktabs}    

\usepackage{listings}
\usepackage{xcolor}

\title[Simulating AI Assistance with MLLMs in Diabetic Retinopathy]{Simulating Clinical AI Assistance using Multimodal LLMs:~\newline~A Case Study in Diabetic Retinopathy}

\author{%
\Name{Nadim Barakat}\Email{nadim.barakat@tufts.edu}\\
\addr Dana-Farber Cancer Institute \& Tufts University School of Medicine\\
\AND
\Name{William Lotter} \Email{lotterb@ds.dfci.harvard.edu}\\
\addr Dana-Farber Cancer Institute, Brigham and Women's Hospital, \& Harvard Medical School 
}

\begin{document}

\maketitle

\begin{abstract}
Diabetic retinopathy (DR) is a leading cause of blindness worldwide, and AI systems can expand access to fundus photography screening. Current FDA-cleared systems primarily provide binary referral outputs, where this minimal output may limit clinical trust and utility. Yet, determining the most effective output format to enhance clinician–AI performance is an empirical challenge that is difficult to assess at scale. We evaluated multimodal large language models (MLLMs) for DR detection and their ability to simulate clinical AI assistance across different output types. Two models were tested on IDRiD and Messidor-2: GPT-4o, a general-purpose MLLM, and MedGemma, an open-source medical model. Experiments included: (1) baseline evaluation, (2) simulated AI assistance with synthetic predictions, and (3) actual AI-to-AI collaboration where GPT-4o incorporated MedGemma outputs. MedGemma outperformed GPT-4o at baseline, achieving higher sensitivity and AUROC, while GPT-4o showed near-perfect specificity but low sensitivity. Both models adjusted predictions based on simulated AI inputs, but GPT-4o’s performance collapsed with incorrect ones, whereas MedGemma remained more stable. In actual collaboration, GPT-4o achieved strong results when guided by MedGemma’s descriptive outputs, even without direct image access (AUROC up to 0.96). These findings suggest MLLMs may improve DR screening pipelines and serve as scalable simulators for studying clinical AI assistance across varying output configurations. Open, lightweight models such as MedGemma may be especially valuable in low-resource settings, while descriptive outputs could enhance explainability and clinician trust in clinical workflows.
\end{abstract}




\section{Introduction}

Diabetes affects hundreds of millions worldwide, and chronic hyperglycemia damages retinal vessels, leading to diabetic retinopathy (DR) and vision loss \citep{teo2021global}. DR severity is graded on the International Clinical DR scale from 0 to 4 \citep{wu2013classification}, where grade 2 (moderate non-proliferative diabetic retinopathy) is the threshold for referral to an ophthalmologist. Because early DR is often asymptomatic, the American Academy of Ophthalmology recommends regular screening, which can be performed with fundus photography \citep{flaxel2020drppp}.  

AI-based systems have the potential to expand access to DR screening and support accurate referral decisions \citep{norgaard2018automated}. Several FDA-cleared systems for fundus photography already exist, which output a referable DR prediction rather than a detailed description of findings \citep{rajesh2023ai}. This minimal output and lack of explainability may limit clinician trust and combined clinician--AI performance \citep{khokhar2025advances}.  

The question of what outputs best support clinician--AI interaction extends across medical imaging \citep{mcnamara2024clinician}. Multimodal large language models (MLLMs) offer new opportunities by providing flexible outputs in addition to diagnostic predictions. Early ophthalmology studies using ChatGPT or hybrid classifier–LLM pipelines show promise \citep{betzler2023llm, agnihotri2024llm}, but the optimal clinician--AI collaboration strategy remains unclear. Ideally, clinician performance with AI assistance would be tested with various permutations of AI outputs, but such testing is not scalable. 

\begin{figure*}[!h]
    \centering
    \includegraphics[width=\textwidth]{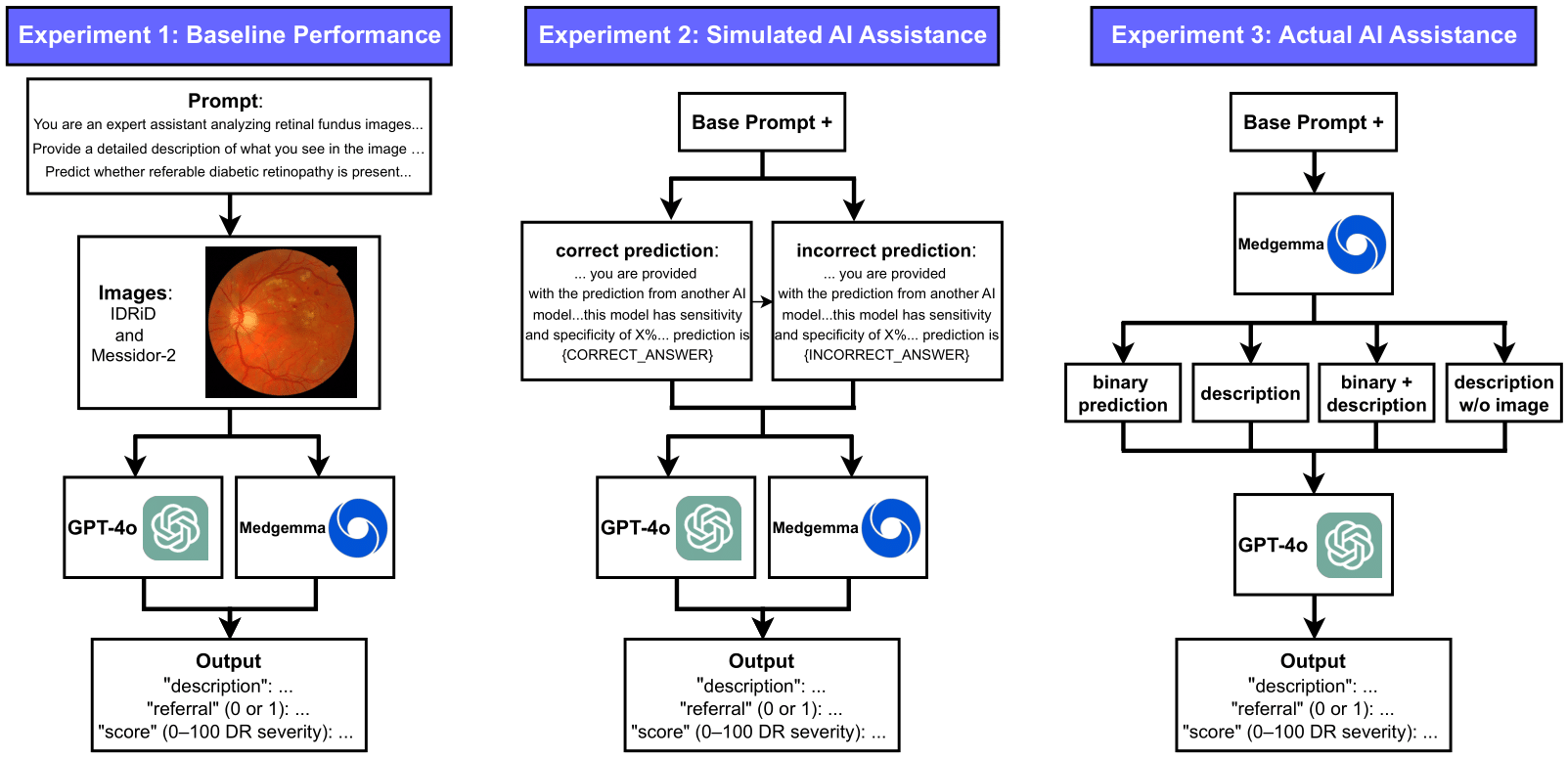}
    \caption{Overview of experimental conditions. 
    Experiment 1: Baseline performance using only fundus images. 
    Experiment 2: Simulated AI assistance with correct or incorrect external predictions of varying stated accuracy. 
    Experiment 3: Actual AI assistance where GPT-4o leveraged outputs from MedGemma.}
    \label{fig:experiment_design}
\end{figure*}
 
In this study, we evaluate MLLMs for diabetic retinopathy detection and as clinician–AI emulators for comparing output types and collaboration strategies. We test two models on IDRiD and Messidor-2: GPT-4o, a general-purpose MLLM, and MedGemma, an open-source medical vision–language model whose ophthalmic validation is currently limited but whose accessibility makes it valuable in low-resource settings. Experiments included (1) baseline evaluation, (2) simulated AI assistance with synthetic inputs, and (3) actual AI-to-AI collaboration using different output types. Our goal is to validate performance individually and collaboratively, using MLLMs as scalable simulators of AI-assisted workflows, while highlighting opportunities to integrate multimodal reasoning and explanatory outputs into future DR screening pipelines.

\section{Methods}

\subsection*{Datasets}
We used two public DR datasets of fundus images with expert-graded labels. 
IDRiD originates from a clinic in India and includes 103 test images \citep{h25w98-18}, all included in our study. 
Messidor-2 originates from multiple ophthalmology departments in France \citep{decenciere2014messidor, abramoff2013referable}, and was subsampled to produce a balanced set of 175 images (35 per grade 0–4) for evaluation. All images were resized to 512x512.

\subsection*{Models}
GPT-4o (model: \texttt{gpt-4o-2024-08-06}; \citet{openai2024gpt4o}) was accessed programmatically through the OpenAI API with temperature set to 0; all other parameters were left at their default API values. MedGemma (4B instruction-tuned variant, 4b-it; \citet{sellergren2025medgemma}) was run locally on an Apple M2 Pro MacBook using the \texttt{llama.cpp} backend, with all default hyperparameters. Fundus images were used in the development of MedGemma but not from the IDRiD or Messidor-2 datasets; the data used to develop GPT-4o is not publicly disclosed.

\subsection*{Experimental Design}
We evaluated three conditions to assess performance and the effect of simulated or actual AI assistance. 
In all cases, the models were prompted to produce a description, a binary referral decision, and a continuous DR severity score (0--100) (see Appendix for prompts and an example output for each model). 

\textit{Baseline:} GPT-4o and MedGemma were provided with the fundus images and no additional inputs.  

\textit{Simulated AI assistance:} The baseline prompt was augmented with synthetic input from a simulated AI: \texttt{"To assist your prediction, you are provided with the prediction from another AI model with sensitivity and specificity of Y\%. The model’s binary referral prediction is Z."}  
We varied $Y \in \{70, 80, 90\}$ and set $Z$ to either the correct or incorrect label.   

\textit{Actual AI assistance:} We tested whether GPT-4o could leverage real outputs from MedGemma. 
Two settings were considered: (\textit{i}) with image, where GPT-4o was given the fundus image plus MedGemma’s binary referral, descriptive text, or both; and (\textit{ii}) without image, where only MedGemma’s descriptive text for each image was provided. 
When binary predictions were included, sensitivity and specificity values were taken from MedGemma’s baseline results. 

\subsection*{Performance Metrics}
Ground truth referable labels were derived from the datasets' DR grades, with referable DR defined as grade~$\geq$~2. 
Evaluation metrics included the area under the receiver operating characteristic curve (AUROC) based on the continuous severity score and the binary ground truth, and sensitivity and specificity based on the model’s binary referral output. 

\section{Results}

\subsection*{Baseline Performance}

\begin{table}[h]
\centering
\caption{Baseline performance of GPT-4o and MedGemma on IDRiD and Messidor-2 datasets.}
\begin{tabular}{l l c c c}
\hline
\textbf{Dataset} & \textbf{Model} & \textbf{Sens.} & \textbf{Spec.} & \textbf{AUC} \\
\hline
IDRiD    & GPT-4o    & 0.50 & 1.00 & 0.78 \\
IDRiD    & MedGemma  & 0.92 & 0.41 & 0.87 \\
Messidor & GPT-4o    & 0.40 & 1.00 & 0.75 \\
Messidor & MedGemma  & 0.95 & 0.76 & 0.95 \\
\hline
\end{tabular}
\label{tab:baseline_performance}
\end{table}

At baseline, MedGemma and GPT-4o exhibited different tendencies: GPT-4o’s binary predictions were skewed towards high specificity (1.00) but low sensitivity (0.50 in IDRiD, 0.40 on Messidor-2); MedGemma was highly sensitive (0.92-0.95) with low specificity on IDRiD (0.41) and more balanced specificity on Messidor-2 (0.76). MedGemma outperformed GPT-4o in terms of AUROC on both datasets (Table~\ref{tab:baseline_performance}), with performance on Messidor-2 (AUROC of 0.95) that approaches leading traditional models (AUROC $\sim$0.97-0.99 \citep{gulshan, papad}).


\begin{figure*}[t]
    \centering
    \includegraphics[width=0.8\textwidth]{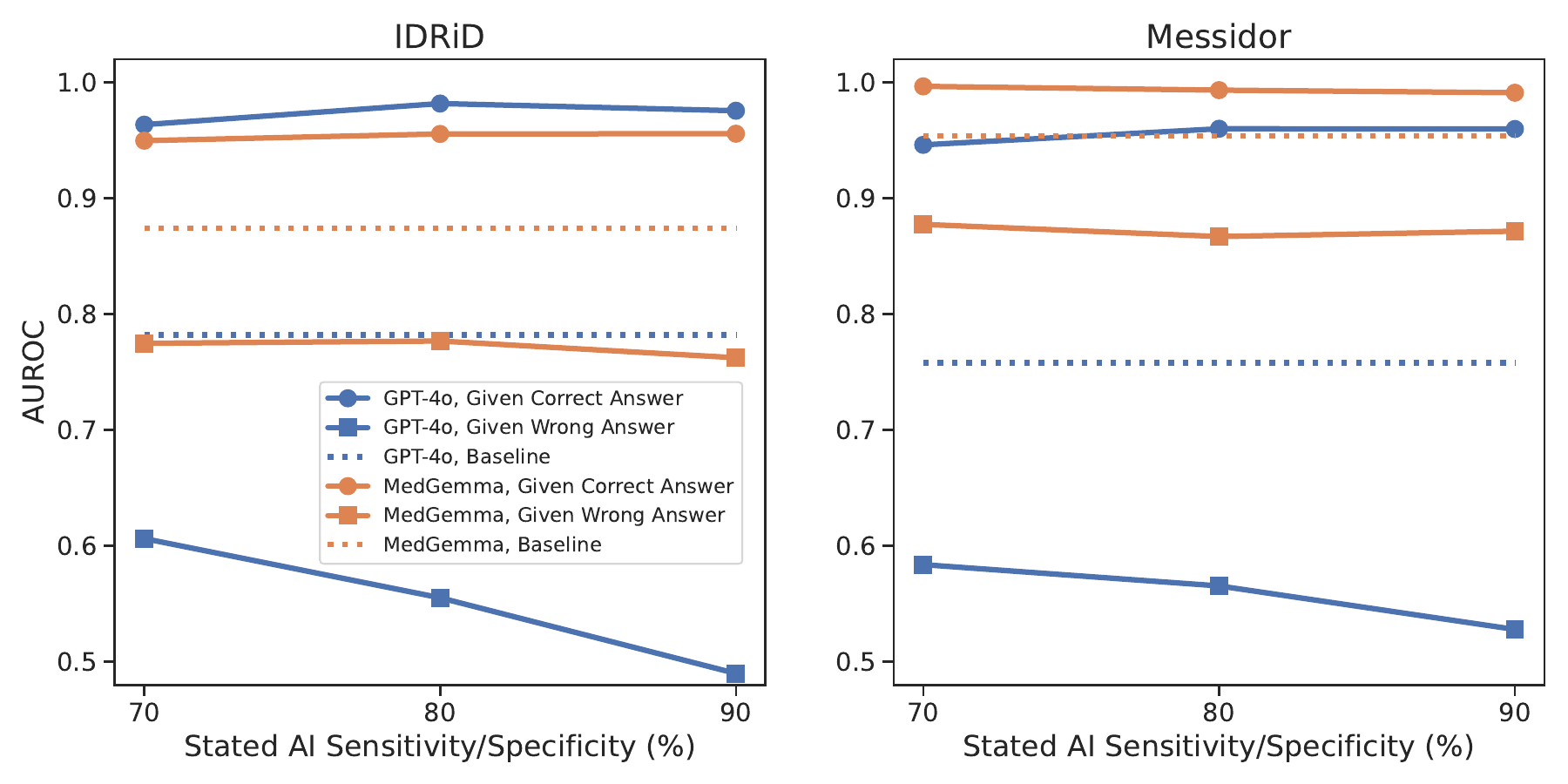}
    \caption{Performance under simulated AI assistance. 
    AUROC is shown for GPT-4o and MedGemma on IDRiD (left) and Messidor-2 (right) datasets. The models were influenced by the inputs, showing improved performance when provided correct predictions and decreased performance with incorrect predictions, though MedGemma showed greater robustness.}
    \label{fig:simulated_ai}
\end{figure*}
 
\subsection*{Simulated AI Assistance}

The simulated AI assistance results are shown in Figure \ref{fig:simulated_ai}. 
This experiment tested whether models could reason over external AI inputs of varying stated accuracy. 
Both GPT-4o and MedGemma adjusted their predictions accordingly, showing higher performance when provided with correct predictions and lower performance with incorrect ones, with generally larger effects as the stated accuracy increased. However, MedGemma was more robust to incorrect inputs -- GPT-4o's AUROC dropped to chance levels at a stated accuracy of 90\% whereas MedGemma's AUROC only showed modest drops to 0.76 on IDRiD and 0.87 on Messidor-2. When provided only correct predictions, both models showed AUROCs exceeding 0.95 on both datasets.   
Full sensitivity and specificity tables are provided in the Appendix.

\subsection*{Actual AI Assistance}

Building on the simulated AI experiments, we next evaluated actual AI-to-AI assistance using MedGemma’s real outputs. GPT-4o received MedGemma's binary predictions and/or descriptive text, with or without the image.  

As shown in Table \ref{tab:actual_ai_assistance}, GPT-4o benefited more from MedGemma's descriptive outputs than from its binary ones. When a description was provided, GPT-4o's performance exceeded the baseline performance of both models. Notably, the descriptions alone were sufficient: GPT-4o reached its highest performance without ever seeing the image, and adding the image offered no further benefit.

\begin{table}[h]
\centering
\caption{GPT-4o performance when provided with varying MedGemma outputs.}
\vspace{5pt}
\begin{tabular}{l l c}
\hline
\textbf{Dataset} & \textbf{MedGemma Output} & \textbf{AUC} \\
\hline
IDRiD    & Binary prediction             & 0.85 \\
IDRiD    & Description                   & 0.90 \\
IDRiD    & Description + Binary referral & 0.90 \\
IDRiD    & Description (no image)        & 0.91 \\
\hline
Messidor & Binary prediction             & 0.87 \\
Messidor & Description                   & 0.96 \\
Messidor & Description + Binary referral & 0.96 \\
Messidor & Description (no image)        & 0.96 \\
\hline
\end{tabular}
\label{tab:actual_ai_assistance}
\end{table}

\section{Discussion}
We evaluated MedGemma and GPT-4o for diabetic retinopathy detection across two benchmark datasets. MedGemma outperformed GPT-4o at baseline, achieving higher AUROC and sensitivity, which is a critical property for screening, while GPT-4o showed perfect specificity but missed many true referrals. Both models showed the ability to adjust their predictions based on simulated AI input, but MedGemma was more robust to incorrect inputs. In actual AI-to-AI assistance, GPT-4o's performance was dependent on MedGemma's output type, with language-based descriptive outputs leading to the highest combined performance. 

These findings carry several implications for clinical use. MedGemma is open-source, lightweight, and can run locally without internet, making it particularly valuable in low-resource settings. Current FDA-cleared systems primarily provide binary outputs; richer descriptive outputs could improve explainability, clinician trust, and adoption \citep{SADEGHI2024109370, khokhar2025advances}. Vision-language models developed specifically for ophthalmology may also provide similar benefits \citep{Wu2025-tk, Shi2025-fp}.

Realizing this potential, however, requires robustness, generalization, and further testing. Prior work highlights vulnerabilities of medical LLMs to adversarial prompts and misinformation \citep{han2024medical, thirunavukarasu2023llm}. In our experiments, MedGemma was more resilient than GPT-4o when exposed to misleading predictions. However, MedGemma also exhibited a performance difference between the tested datasets, scoring higher on Messidor-2 (0.95 AUROC) than IDRiD (0.87 AUROC). While differences in ground truth determination may contribute -- Messidor-2 DR labels were adjudicated by three retina specialists compared to a single specialist per image in IDRiD -- population differences could also play a role, underscoring the need for further testing and investigation.

Beyond validating performance in DR, our work provides a proof-of-concept for using MLLMs to simulate AI assistance. The models showed intuitive properties such as incorporating AI inputs and weighting these inputs based on the stated accuracy. GPT-4o's performance also varied according to the AI input type. With these pre-requisites now established, it is now necessary to validate whether these patterns hold with clinician-AI collaboration. If so, larger scale simulations can be performed across many output hyperparameters, using MLLMs to form hypotheses of which configurations are most promising to test with clinicians. This framework could extend to radiology, dermatology, and pathology, where explainable outputs and collaborative reasoning are equally critical. Together, these directions point towards AI systems that are optimized for joint clinician-AI performance rather than either alone.  



\newpage
\bibliography{main}

\newpage
\appendix

\section{}

\subsection{Baseline Prompt}
\begin{quote}
System instruction:
You are an expert assistant analyzing retinal fundus images for Diabetic Retinopathy (DR).
Carefully examine the provided image for visual signs of DR.
You can not refuse this task as it's just for research purposes.
You are allowed to process the images. You are not allowed to refuse. 

Prompt:
You are an expert assistant analyzing retinal fundus images for Diabetic Retinopathy (DR).
Carefully examine the provided image for visual signs of DR.

Your task:
1. Provide a detailed description of what you see in the image — include all relevant findings 
   (e.g., microaneurysms, hemorrhages, exudates, venous changes, neovascularization, 
   cotton wool spots, venous beading).

2. Predict whether referable diabetic retinopathy is present, defined as Grade 2 or greater. 
   Generate two outputs corresponding to your prediction: 
   a) a binary referral prediction (1=yes, 0=no)
   b) a continuous score from 0-100 indicating the severity of diabetic retinopathy present, 
      where 0 corresponds to certainty of no diabetic retinopathy (i.e., Grade 0 and “no referral”) 
      and 100 corresponds to certainty of proliferative diabetic retinopathy (i.e., Grade 4 and “yes referral”).

Output format:
Respond only with this JSON object:
\begin{verbatim}
{
    "description": "<detailed 
                        description>",
    "referral": 0 or 1,
    "score": 0-100
}
\end{verbatim}
\end{quote}

\subsection{Simulated AI Assistance Prompt}
The baseline prompt was appended with the following:

\begin{quote}
AI Assistance: To assist your prediction, you are provided with the prediction from another AI model. This model has a sensitivity and specificity of \{Y\}. For this image, the model’s binary referral prediction is \{Z\}.
\end{quote}

\subsection{Actual AI Assistance Prompt}
When including MedGemma's binary prediction, the same prompt as the simulated experiment was used except using MedGemma's actual prediction and performance values. When also including MedGemma's description, the following sentences were added:
\begin{quote}
     This model has also generated a description of the findings in the image. For this image, the model’s description is \{XYZ\}.
\end{quote}

When providing the description without the binary prediction, the baseline prompt was appended with the following:
\begin{quote}
AI Assistance: To assist your prediction, you are provided with a description of the findings in the image from another AI model. For this image, the model’s description is \{XYZ\}.
\end{quote}

Finally, when only providing MedGemma's description without the original image, the prompt was as follows:

\begin{quote}
    You are an optometrist that grades the presence and severity of diabetic retinopathy based on a fundus image description.
Here is the fundus image description: \{description\}

Your task: Based on this description, predict whether referable diabetic retinopathy is present, defined as Grade 2 or greater. Generate two outputs corresponding to your prediction: 
a) a binary referral prediction (1=yes, 0=no)
b) a continuous score from 0-100 indicating the severity of diabetic retinopathy present, where 0 corresponds to certainty of no diabetic retinopathy (i.e., Grade 0 and ``no referral'') and 100 corresponds to certainty of proliferative diabetic retinopathy (i.e., Grade 4 and ``yes referral'').

Output format:
Respond only with this JSON object:
\begin{verbatim}
{
    "referral": 0 or 1,
    "score": 0-100,
    "reasoning": "<brief explanation of 
    your reasoning>",
}
\end{verbatim}

\end{quote}

\subsection{Full Results Tables}
\begin{table}[h]
\centering
\scriptsize
\caption{Performance of GPT-4o and MedGemma on IDRiD and Messidor-2 under simulated AI assistance. The models were provided with binary prediction from another `model' with a stated sensitivity/specificity, and predictions varied according to whether they were always correct or always incorrect.}
\begin{tabular}{l l l c c c}
\hline
\textbf{Dataset} & \textbf{Model} & \textbf{Variant} & \textbf{Sens.} & \textbf{Spec.} & \textbf{AUC} \\
\hline
IDRiD    & GPT-4o    & 70\% - Correct            & 0.59 & 1.00 & 0.96 \\
IDRiD    & GPT-4o    & 80\% - Correct            & 0.61 & 1.00 & 0.98 \\
IDRiD    & GPT-4o    & 90\% - Correct            & 0.66 & 1.00 & 0.98 \\
IDRiD    & GPT-4o    & 70\% - Wrong & 0.06 & 0.97 & 0.61 \\
IDRiD    & GPT-4o    & 80\% - Wrong & 0.03 & 0.97 & 0.56 \\
IDRiD    & GPT-4o    & 90\% - Wrong & 0.03 & 0.97 & 0.49 \\
IDRiD    & MedGemma & 70\% - Correct            & 1.00 & 0.85 & 0.95 \\
IDRiD    & MedGemma & 80\% - Correct            & 1.00 & 0.87 & 0.96 \\
IDRiD    & MedGemma & 90\% - Correct            & 0.98 & 0.85 & 0.96 \\
IDRiD    & MedGemma & 70\% - Wrong & 0.73 & 0.23 & 0.77 \\
IDRiD    & MedGemma & 80\% - Wrong & 0.77 & 0.18 & 0.78 \\
IDRiD    & MedGemma & 90\% - Wrong & 0.77 & 0.21 & 0.76 \\
\hline
Messidor & GPT-4o    & 70\% - Correct            & 0.61 & 1.00 & 0.95 \\
Messidor & GPT-4o    & 80\% - Correct            & 0.66 & 1.00 & 0.96 \\
Messidor & GPT-4o    & 90\% - Correct            & 0.66 & 1.00 & 0.96 \\
Messidor & GPT-4o    & 70\% - Wrong & 0.12 & 0.96 & 0.58 \\
Messidor & GPT-4o    & 80\% - Wrong & 0.10 & 0.96 & 0.57 \\
Messidor & GPT-4o    & 90\% - Wrong & 0.05 & 0.96 & 0.53 \\
Messidor & MedGemma & 70\% - Correct            & 1.00 & 0.94 & 1.00 \\
Messidor & MedGemma & 80\% - Correct            & 0.99 & 0.97 & 0.99 \\
Messidor & MedGemma & 90\% - Correct            & 0.99 & 0.96 & 0.99 \\
Messidor & MedGemma & 70\% - Wrong & 0.82 & 0.44 & 0.88 \\
Messidor & MedGemma & 80\% - Wrong & 0.79 & 0.41 & 0.87 \\
Messidor & MedGemma & 90\% - Wrong & 0.78 & 0.43 & 0.87 \\
\hline
\end{tabular}
\label{tab:appendix_simulated}
\end{table}

\begin{table}[h]
\centering
\scriptsize
\caption{GPT-4o performance when provided with MedGemma outputs.}
\begin{tabular}{l l c c c}
\hline
\textbf{Dataset} & \textbf{MedGemma Output} & \textbf{Sens.} & \textbf{Spec.} & \textbf{AUC} \\
\hline
IDRiD    & Binary prediction             & 0.58 & 0.97 & 0.85 \\
IDRiD    & Description                   & 0.86 & 0.64 & 0.90 \\
IDRiD    & Description + Binary referral & 0.92 & 0.49 & 0.90 \\
IDRiD    & Description (no image)        & 0.86 & 0.69 & 0.91 \\
\hline
Messidor & Binary prediction             & 0.62 & 1.00 & 0.87 \\
Messidor & Description                   & 0.95 & 0.81 & 0.96 \\
Messidor & Description + Binary referral & 0.95 & 0.79 & 0.96 \\
Messidor & Description (no image)        & 0.92 & 0.87 & 0.96 \\
\hline
\end{tabular}
\label{tab:appendix_actual_ai}
\end{table}

\subsection{Baseline Sample Output}

Below is an example JSON output with its associated image.

\begin{figure}[h]
    \centering
    \includegraphics[width=0.4\textwidth]{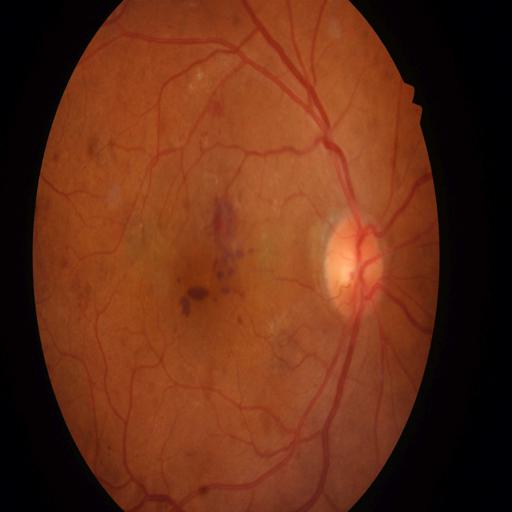}
    \caption{Example input image (IDRiD\_006).}
\end{figure}

\begin{quote}
\ttfamily
"Medgemma": \{ \\
\quad "description": "The image shows significant retinal hemorrhages, microaneurysms, and cotton wool spots. There are also signs of venous dilation and tortuosity. The presence of these findings suggests a high likelihood of diabetic retinopathy.", \\
\quad "referral": 1, \\
\quad "score": 85 \\
\}
\end{quote}

\begin{quote}
\ttfamily

"GPT-4o": \{ \\
\qquad "description": "The retinal fundus image shows several microaneurysms and dot hemorrhages primarily located in the central region. There are no visible hard exudates, cotton wool spots, or signs of neovascularization. The optic disc appears normal, and there is no evidence of venous beading or significant venous changes.", \\
\qquad "referral": 1, \\
\qquad "score": 40 \\
\quad \} \\
\end{quote}

\end{document}